\documentclass[10pt,twocolumn,letterpaper]{article}

\usepackage[pagenumbers]{cvpr} 

\usepackage{graphicx}
\usepackage{amsmath}
\usepackage{amssymb}
\usepackage{booktabs}

\usepackage{subcaption}
\usepackage{xcolor}
\usepackage{color}
\usepackage{soul}
\usepackage{amsmath, amsthm, amssymb}
\usepackage{algorithm,algorithmic,paralist}
\usepackage{bm}
\usepackage[american]{babel}
\usepackage{microtype}
\usepackage{latexsym}
\usepackage{booktabs}
\usepackage{multirow}
\usepackage{array}

\usepackage[pagebackref,breaklinks,colorlinks]{hyperref}

\usepackage[capitalize]{cleveref}
\crefname{section}{Sec.}{Secs.}
\Crefname{section}{Section}{Sections}
\Crefname{table}{Table}{Tables}
\crefname{table}{Tab.}{Tabs.}

\def\cvprPaperID{6978} 
\def\confName{CVPR}
\def\confYear{2022}

\begin{document}

\title{Active Learning for Open-set Annotation}

\author{Kun-Peng Ning\\
{\tt\small ningkp@nuaa.edu.cn}
\and
Xun Zhao\\
{\tt\small emmaxunzhao@gmail.com}
\and
Yu Li\\
{\tt\small yul@illinois.edu}
\and
Sheng-Jun Huang$^*$\\
{\tt\small huangsj@nuaa.edu.cn}
}
\maketitle

\begin{abstract}

	Existing active learning studies typically work in the closed-set setting by assuming that all data examples to be labeled are drawn from known classes.
	However, in real annotation tasks, the unlabeled data usually contains a large amount of examples from unknown classes, resulting in the failure of most active learning methods.
	To tackle this open-set annotation (OSA) problem, we propose a new active learning framework called LfOSA, which boosts the classification performance with an effective sampling strategy to precisely detect examples from known classes for annotation.
	The LfOSA framework introduces an auxiliary network to model the per-example max activation value (MAV) distribution with a Gaussian Mixture Model, which can dynamically select the examples with highest probability from known classes in the unlabeled set.
	Moreover, by reducing the temperature $T$ of the loss function, the detection model will be further optimized by exploiting both known and unknown supervision.
	The experimental results show that the proposed method can significantly improve the selection quality of known classes, and achieve higher classification accuracy with lower annotation cost than state-of-the-art active learning methods.
	To the best of our knowledge, this is the first work of active learning for open-set annotation.
	

\end{abstract}

\section{Introduction}
The remarkable success of deep learning is largely attributed to the collection of large datasets with human annotated labels~\cite{lecun2015deep,huangasynchronous}. Nevertheless, it is extremely expensive and time-consuming to label large scale data with high-quality annotations~\cite{settles2009active,tang2021qbox}. It is thus a significant challenge to learn with limited labeled data.

Active learning (AL) is a primary approach to tackle this problem. It iteratively selects the most useful examples from the unlabeled data to query their labels from the oracle, achieving competitive performance while reducing annotation costs~\cite{settles2009active,sun2010survey,huang2018cost}. 
Existing AL methods typically work in a closed-set setting where the labeled and unlabeled data are both drawn from the same class distribution.

\begin{figure}[!tb]
	\begin{center}

		\begin{subfigure}{1\linewidth}
			\centering
			\includegraphics[width=1\textwidth]{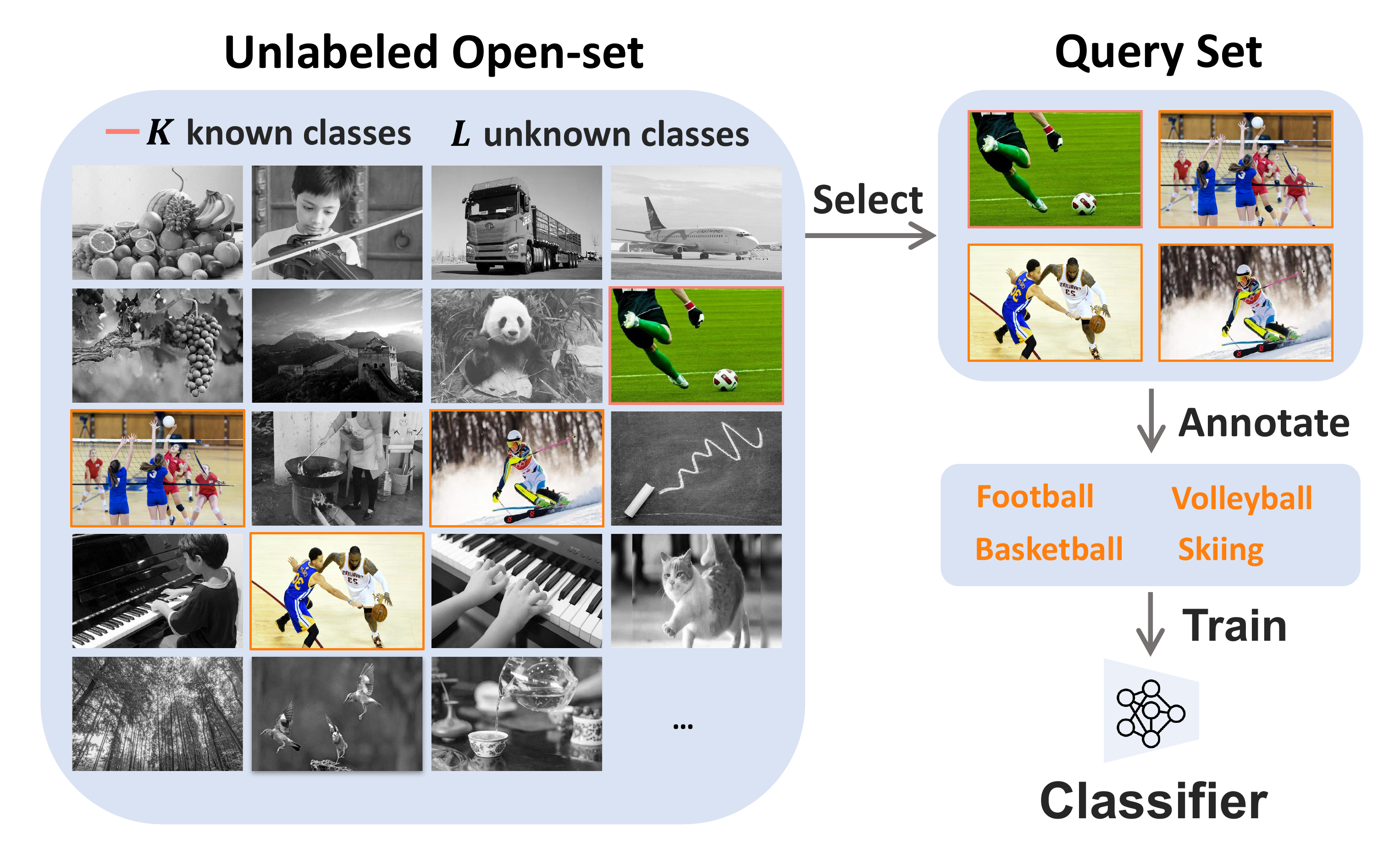}
		\end{subfigure}

		\caption{The illustration of open-set annotation (OSA) problem. The unlabeled open-set contains $K$ known classes (color images with border) and $L$ unknown classes (gray-scale images without border). The goal is to find and annotate the examples from known classes for training the classifier.}
		\vspace{-0.2cm}
		\label{fig.osa}
	\end{center}
\end{figure}

    However, in some real world scenarios, the unlabeled data are usually uncontrolled and large amounts of data examples are from unknown classes. Figure~\ref{fig.osa} shows an example to train a new model for sports image classification for image-sharing social platform where the database contains trillions of images from unconstrained categories uploaded by the users. A large number of images in the unlabeled open-pool are actually from irrelevant classes (\eg cats and pianos \etc). These images are usually useless for learning the sports classifier, so that it will be a waste of annotation budget to select them to label. On the other hand, existing closed-set AL system cannot precisely distinguish these irrelevant images from unknown classes but tends to choose them for annotation as they contain more uncertainty or information. Therefore an effective and practical AL system in the real-world open-set case is highly desired which can 1) precisely distinguish the examples of unwanted classes and 2) meanwhile query the most useful cases from the wanted classes to train the classifier. 
In this paper, we formulate this problem as an open-set annotation (OSA) task. As shown in Figure~\ref{fig.osa}, the unlabeled set contains $K$ known classes and $L$ unknown classes, where $L \textgreater K$. 
The goal is to precisely filter out examples from unknown classes, while actively select a query set that contains examples from known classes as pure as possible.
To overcome this challenge, we propose a new active learning framework called LfOSA (\textbf{L}earning \textbf{f}rom \textbf{O}pen-\textbf{S}et \textbf{A}nnotation), which includes two networks for detection and classification respectively.
Specifically, the detector models the per-example max activation value (MAV) distribution with a Gaussian Mixture Model~\cite{reynolds2009gaussian} to dynamically divide the unlabeled open-set into known and unknown set, 
then examples from the known set with larger certainty will be selected to construct a query set for annotation.
After labeling, the classification model will be updated with the new examples from known classes.
Meanwhile, as the query set will inevitably include a few invalid examples from unknown classes, these invalid examples will be utilized as negative training examples to update the detector, such that the detector can maintain a higher recall to identify known-class examples from the unlabeled open-set.
Moreover, by reducing the temperature $T$ of the cross-entropy (CE) loss, the distinguishability of the detector is further enhanced.

Experiments are conducted on multiple datasets with different mismatch ratios of known and unknown classes. The experimental results demonstrate that the proposed approach can significantly improve the selection quality of known classes, and achieve higher classification accuracy with lower annotation cost than state-of-the-art active learning methods.

The major contributions can be summarized as follows:
\begin{itemize}
	\item We formalize a new problem of open-set annotation (OSA) for real-world large-scale annotation tasks. It raises a practical challenge on how to maintain a higher recall to find the examples of known classes from a large unlabeled open-set for target model training.
	\item We propose a new active learning framework LfOSA to address the OSA problem. It iteratively trains two networks, one for distinguishing the known and unknown classes, while the other one for classification of target task. To the best of our knowledge, this is the first work on active learning for open-set annotation.
	\item The experimental results validate that the proposed approach can significantly improve the selection precision and recall, while achieving higher classification accuracy with lower annotation cost than state-of-the-art active learning methods.
\end{itemize}


\section{Related Work}
\textbf{Active learning.}\quad Active learning as a large-scale annotation tool has achieved a great success for learning with limited labeled data~\cite{hua2008online,qian2013fast}. It reduces the labeling cost by selecting the most useful examples to query their labels. Most researches focus on designing effective sampling strategies to make sure that the selected examples can improve the model performance most~\cite{fu2013survey}. During the past decades, many criteria have been proposed for selecting examples~\cite{fu2013survey,huang2010active,lewis1994sequential,seung1992query,ning2021improving,you2014diverse,geman1992neural,roy2001toward}. Among of these approaches, some of them prefer to select the most informative examples to reduce the model uncertainty~\cite{lewis1994sequential,seung1992query,you2014diverse}, while some others prefer to select the most representative examples to match the data distribution~\cite{geman1992neural,roy2001toward}. Moreover, some studies try to combine informativeness and representativeness to achieve better performance~\cite{huang2013active,huang2010active}. These standard active learning methods are usually based on the closed-set assumption that the unlabeled data are drawn from known classes, which leads to failure on the open-set annotation (OSA) task. 

\textbf{Open-set recognition.} Open-set recognition (OSR) attempts to address the classification setting where inference can face examples from unseen classes during training~\cite{scheirer2012toward,scheirer2014probability,jain2014multi}. Its goal is to learn an open-set classifier with a mechanism to reject such unknown examples~\cite{geng2020recent}.
A representative approach called OpenMax has achieved remarkable results on the OSR problem, which employs deep neural networks to OSR by combining Extreme Value Theory with neural networks~\cite{bendale2016towards}. Another follow-up work proposed G-OpenMax by adopting GAN~\cite{creswell2018generative,goodfellow2014generative} for generating examples which are highly similar to training examples yet do not belong any of the training classes~\cite{ge2017generative}. However, these OSR methods usually fail on the OSA problem for the following two essential differences between both.
First, the training process of OSR has abundant labeled data and is based on the closed-set assumption, while the OSA problem has limited labeled data and its unlabeled data are open set.
Second, the OSR focuses on rejecting unknown examples in testing phase after training, while the OSA aims to find more known examples from the unlabeled open-set for target model optimizing in training phase.
\begin{figure*}[!th]
	\begin{center}

		\begin{subfigure}{1\linewidth}
			\centering
			\includegraphics[width=1\textwidth]{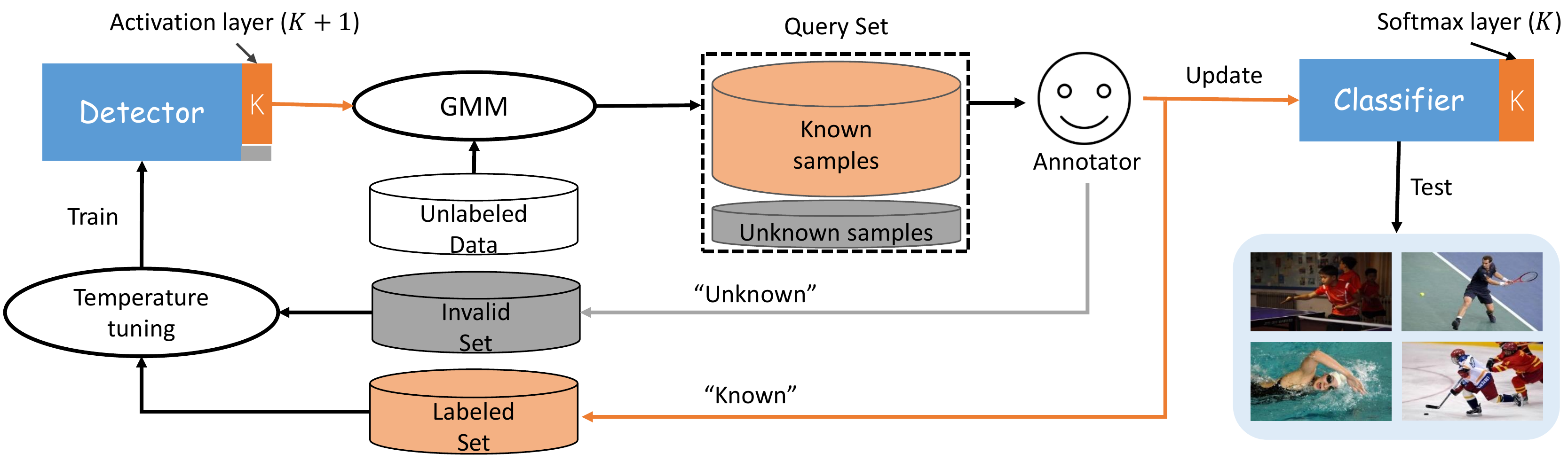}
		\end{subfigure}
		\caption{The framework of LfOSA. It includes two networks for detection and classification. The detector attempts to construct a query set for annotation by GMM modeling. After labeling, two networks will be updated for next iteration.}
		
		\vspace{-0.3cm}
		\label{fig.framework}
	\end{center}
\end{figure*}

\section{The Proposed Approach}
In this section, we first formalize the open-set annotation (OSA) problem, and then introduce the proposed LfOSA approach in detail.
\subsection{The OSA Problem Setting}
In the OSA problems, we consider a large-scale annotation scenario with a limited labeled set $D_L$ and a huge number of unlabeled open-set $D_U$, where $D_L=\{(x_i^L,y_i^L)\}_{i=1}^{n^L}$ and $D_U = \{x_j^U\}_{j=1}^{n^U}$. Let $D_U=X_{kno} \cup X_{unk}$ and $X_{kno} \cap X_{unk}=\emptyset$, where $X_{kno}$ and $X_{unk}$ denote the examples from known and unknown classes respectively. Each labeled example $x_i^L$ belongs to one of $K$ known classes $Y=\{y_k\}_{k=1}^K$, while an unlabeled example $x_j^U$ may belong to an unknown class not belonging to $Y$. Let $X^{query}$ denotes the query set during each iteration, which consists of unknown query set $X_{unk}^{query}$ and known query set $X_{kno}^{query}$, \ie, $X^{query}=X_{kno}^{query} \cup X_{unk}^{query}$. The goal is to selectively construct the query set that contains known examples as many as possible.

Active learning (AL) iteratively selects the most useful examples from the unlabeled dataset to query their labels from the oracle~\cite{settles2009active}. After annotating the newly selected data, the model can be updated to achieve better performance. Specifically, in the $i$-th iteration, we train a classifier $f_{\theta_C}$ with parameters $\theta_C$ on labeled set $D_L$. Then, a batch of $b$ examples $X^{query}$ are selected with a specific criterion based on the current trained model. After querying their labels, $k_i$ known examples $X_{kno}^{query}$ are annotated and the labeled set is updated to $D_L=D_L \cup X_{kno}^{query}$, while $l_i$ examples $X_{unk}^{query}$ with unknown classes are added to the invalid set $D_I$, where $b=k_i+l_i$. Thus, the recall and precision of known classes in the $i$-th selection can be defined as follow,
\begin{equation}\label{recall}
recall_i = \frac{\sum_{j=0}^{i}k^i}{n_{kno}}
\end{equation}
\begin{equation}\label{precision}
precision_i = \frac{k_i}{k_i+l_i}
\end{equation}
where $n_{kno}$ denotes the number of examples from known classes in the unlabeled set. $recall_i$ calculates how many known examples are queried after $i$ queries, and $precision_i$ denotes the proportion of the target examples in the $i$-th query. Obviously, if we maintain a high precision and recall to accurately select known examples, the trained target classifier will be more effective.

As discussed in the Introduction, most of the traditional AL methods are less effective in OSA problem, because their selection strategies tend to select open-set (unknown) examples with larger uncertainty. These examples from unknown classes are useless for training the target model, and thus traditional AL methods will probably fail with serious waste of the annotation budget.
Fortunately, we should be aware that although these examples are useless for the target model, they could be exploited to improve the detector model for filtering out unknown classes from the open-set data.
Moreover, we find that the activation (penultimate) layer of network has strong ability to distinguish unknown classes based on the observation that the maximum activation value (MAV) of open set examples are often far awary from the average MAV of closed set examples.
By decoupling detection and classification, we propose to exploit examples of both known and unknown classes to train a detector with strong distinguishability and train a classifier for the target task.

\subsection{Algorithm Detail}
The framework of LfOSA is demonstrated in Figure~\ref{fig.framework}, which mainly composed of three components: detector training, active sampling and classifier training. Specifically, we first train a network for detecting unknown examples by exploiting both known and unknown supervision while using a low-temperature mechanism. Then, by modeling per-example max activation value (MAV) distribution with a Gaussian Mixture Model (GMM), the most certain known examples can be actively selected for annotation. Finally, the classification model will be updated with the new examples from known classes. In the following part of this section, we will introduce these three components in detail.

%
\textbf{Detector training.} In addition to classifying $K$ known classes, the detector has been extended with an additional $(K+1)$-th output to predict unknown class. For a given example $x$ from labeled or invalid set, we encode its label $c$ with onehot $p$, \ie, the value of $p_c$ is set to 1 and the others to 0. Then, we train the detector with the following cross-entropy loss:
\begin{equation}\label{eq:loss_r}
\mathcal{L}_D(x,c) = -\sum_{c=1}^{K+1}p_c*log(q_c^T)
\end{equation}
where
\begin{equation*}
q_c^T=\frac{exp(a_c/T)}{\sum_j exp(a_j/T)}.
\end{equation*}
where $a_c$ is the $c$-th activation value of the last fully-connected layer, $T$ is a temperature, which is set with a lower value $(T=0.5)$ to produce a sharper probability distribution $q_c^T$ over classes. Obviously, by minimizing the loss function, examples of known classes will have larger activation values on the first $K$ dimensions and smaller activation values on the $(K+1)$-th dimension, while examples of unknown classes have the opposite phenomenon. Moreover, we find that the distinguishability of the activation layer can be further enhanced by reducing the temperature $T$ of the loss function. A brief analysis is as follows:
\begin{equation}
\frac{\partial \mathcal{L}_D}{\partial a_c}=\frac{1}{T}(q_c^T-p_c)=\frac{1}{T}(\frac{exp(a_c/T)}{\sum_j exp(a_j/T)}-p_c).
\end{equation}
When we reduce the temperature $(T \downarrow)$ of the loss function $\mathcal{L}_R$, the probability distribution $q_c^T$ will be more sharper, thus we have:
\begin{equation*}
T \downarrow \Rightarrow \frac{1}{T} \uparrow, \frac{exp(a_c/T)}{\sum_j exp(a_j/T)}-p_c \uparrow \Rightarrow \frac{\partial \mathcal{L}_D}{\partial a_c} \uparrow.
\end{equation*}
As $\frac{\partial \mathcal{L}_R}{\partial a_c}$ becomes larger, the examples of known and unknown classes will be more distinguishable for the activation value $a_c$.

\begin{algorithm}[tb]
	\caption{The LfOSA algorithm}
	\label{alg:LfOSA}
	\begin{algorithmic}[1]
		\STATE \textbf{Input:}
		\STATE \quad Current detector $f_{\theta_D}$ and classifier $f_{\theta_C}$
		\STATE \quad Current labeled set $D_L$ and invalid set $D_I$
		\STATE \quad Query batch size $b$ and temperature $T$
		\STATE \textbf{Process:}
		\STATE \quad \emph{\textbf{\# Recognizer training}}
		\STATE \quad Update $\theta_D$ by minimizing $\mathcal{L}_D$ in Eq.~\ref{eq:loss_r} from $D_L$ and $D_I$
		\STATE \quad \emph{\textbf{\# Examples sampling}}
		\STATE \quad Inference $mav_i^c$ from detector $\theta_D$ for each unlabeled example $x_i$ as Eq.~\ref{eq:mav}
		\STATE \quad \textbf{while} $c=1,2,...,K$ do
		
		\STATE \quad \quad \emph{\textcolor[rgb]{0.5,0.5,0.5}{\# Collect the MAV set for each prediction class $c$}}
		\STATE \quad \quad $mav^c=\{mav_i^c|f_{\theta_D}(x_i)=c,\forall x_i\in D_U\}$
		\STATE \quad \quad \emph{\textcolor[rgb]{0.5,0.5,0.5}{\# Obtain known probability by GMM}}
		\STATE \quad \quad $\mathcal{W}^c=GMM(mav^c,\theta_D)$
		\STATE \quad \textbf{end}
		\STATE \quad \emph{\textcolor[rgb]{0.5,0.5,0.5}{\# Merge and sort the probability sets of all classes}}
		\STATE \quad $\mathcal{W}=sort(\mathcal{W}^1\cup \mathcal{W}^1\cup ... \cup\mathcal{W}^K)$
		\STATE \quad \emph{\textcolor[rgb]{0.5,0.5,0.5}{\# Obtain the query set }}
		\STATE \quad $X^{query}=\{(x_i, w_i)|w_i \geq \tau, \forall (x_i, w_i)\in (D_U, \mathcal{W})\}$
		\STATE \quad \emph{\textcolor[rgb]{0.5,0.5,0.5}{\# Ask for annotation from Oracle }}
		\STATE \quad Query their labels and obtain $X^{query}_{kno}$ and $X^{query}_{unk}$
		\STATE \quad \emph{\textcolor[rgb]{0.5,0.5,0.5}{\# Update labeled and invalid sets}}
		\STATE \quad $D_L = D_L \cup X^{query}_{kno}, D_I = D_I \cup X^{query}_{unk}$
		\STATE \quad \emph{\textbf{\# Classifier training}}
		\STATE \quad Update $\theta_C$ by minimizing $\mathcal{L}_C$ in Eq.~\ref{eq:loss_c} from $D_L$
		
		\STATE \textbf{Output:} $\theta_D $, $\theta_C$, $D_L$ and $D_I$ for next iteration.
		
	\end{algorithmic}
\end{algorithm}

\textbf{Active sampling.} As mentioned earlier, the goal of OSA task is to precisely select as many known-class examples as possible from the unlabeled open-set. After training the detector as shown above, we find that the activation (penultimate) layer of network has the ability to distinguish unknown examples, that is, the maximum activation value (MAV) of unknown-class examples are often significantly different from the average MAV of known-class examples. Formally, for each unlabeled example $x_i$ with predicted class $c$, its maximum activation value $mav_i^c$ can be defined as follow:
\begin{equation}\label{eq:mav}
mav_i^c = \max_c a^i_c
\end{equation}
All unlabeled examples will be classified into $K+1$ classes according to the prediction of the current detector. We can select the examples predicted as the first $K$ known classes for the next process while filtering out the examples predicted as ``unknown''.
Then, for each known class $c$, we fit a two-component GMM to $mav^c$ using the Expectation-Maximization algorithm, where $mav^c$ is a set of activation values with prediction class $c$.
\begin{equation}
\mathcal{W}^c=GMM(mav^c,\theta_D),
\end{equation}
where $\mathcal{W}^c$ is the probabilities of class $c$. For each unlabeled example $x_i$ from class $c$, its known probability $w_i\in \mathcal{W}^c$ is the posterior probability $p(g|mav_i)$, where $g$ is the Gaussian component with larger mean (larger activation value). Then we merge and sort the probabilities of all categories,
\begin{equation}
\mathcal{W}=sort(\mathcal{W}^1 \cup \mathcal{W}^2 \cup ... \cup \mathcal{W}^K).
\end{equation}
Next, we select the first $b$ examples with highest probability as the query set to ask for annotation. In other words, we can obtain the query set $X^{query}$ by setting a threshold $\tau$ on $w_i$, where $\tau$ is equal to the $b$-th largest known probability:
\begin{equation}
X^{query}=\{(x_i, w_i)|w_i \geq \tau, \forall (x_i, w_i)\in (D_U, \mathcal{W})\}.
\end{equation}
After querying their labels, the labeled and unknown sets will be updated by adding $X_{kno}^{query}$ and $X_{unk}^{query}$, respectively.

\begin{figure*}[!ht]
	\begin{center}
		
		\begin{subfigure}{0.33\linewidth}
			\centering
			\includegraphics[width=1\textwidth]{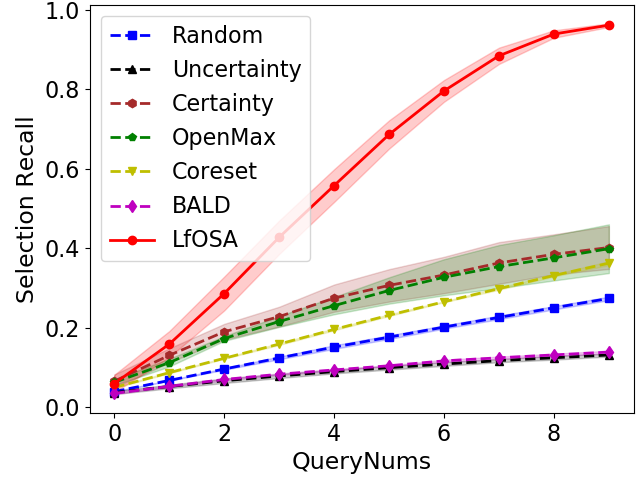}\\
		\end{subfigure}
		\begin{subfigure}{0.33\linewidth}
			\centering
			\includegraphics[width=1\textwidth]{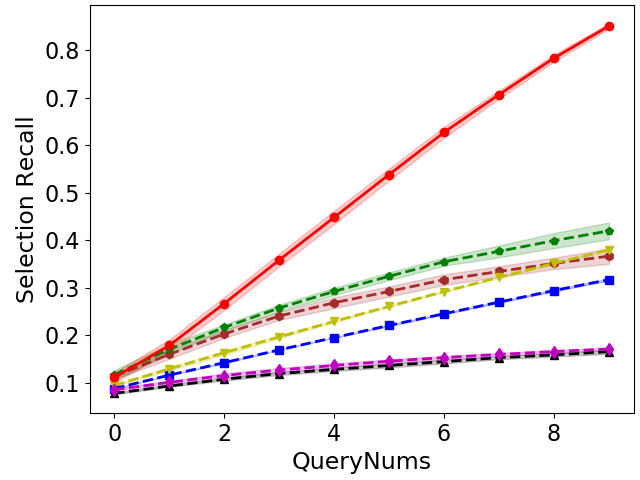}\\
		\end{subfigure}
		\begin{subfigure}{0.33\linewidth}
			\centering
			\includegraphics[width=1\textwidth]{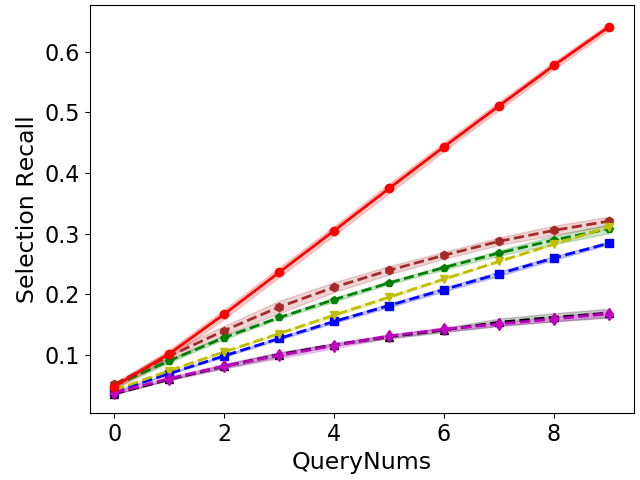}\\
		\end{subfigure}
		
		\begin{subfigure}{0.33\linewidth}
			\centering
			\includegraphics[width=1\textwidth]{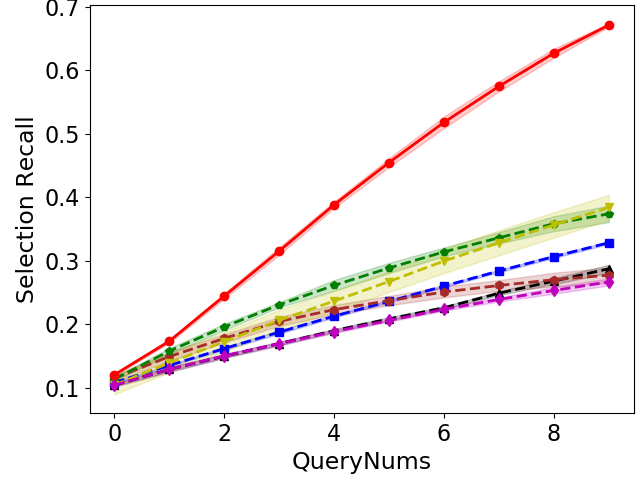}\\
		\end{subfigure}
		\begin{subfigure}{0.33\linewidth}
			\centering
			\includegraphics[width=1\textwidth]{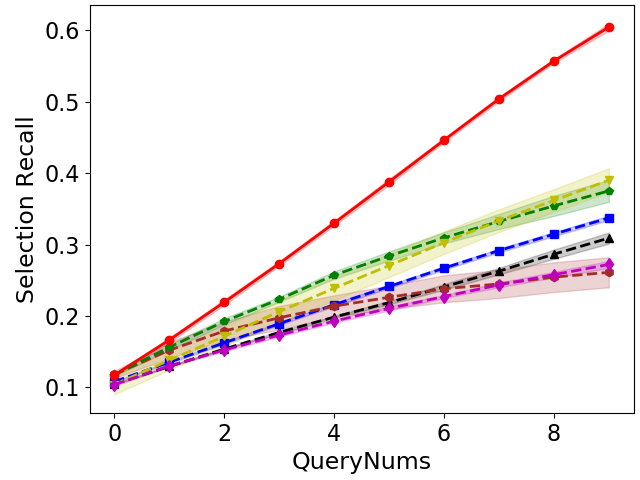}\\
		\end{subfigure}
		\begin{subfigure}{0.33\linewidth}
			\centering
			\includegraphics[width=1\textwidth]{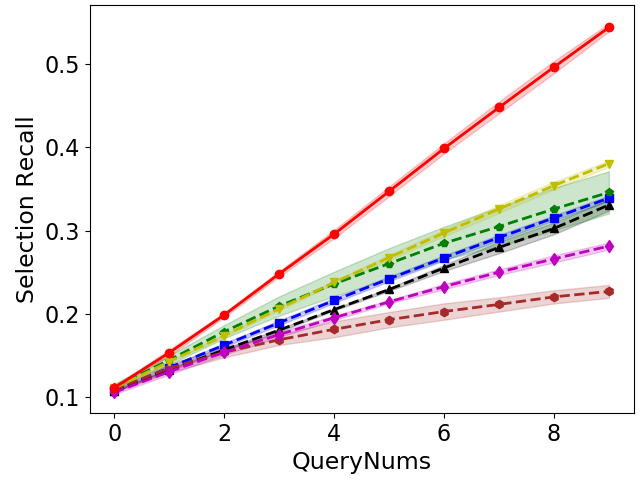}\\
		\end{subfigure}
		
		\begin{subfigure}{0.33\linewidth}
			\centering
			\includegraphics[width=1\textwidth]{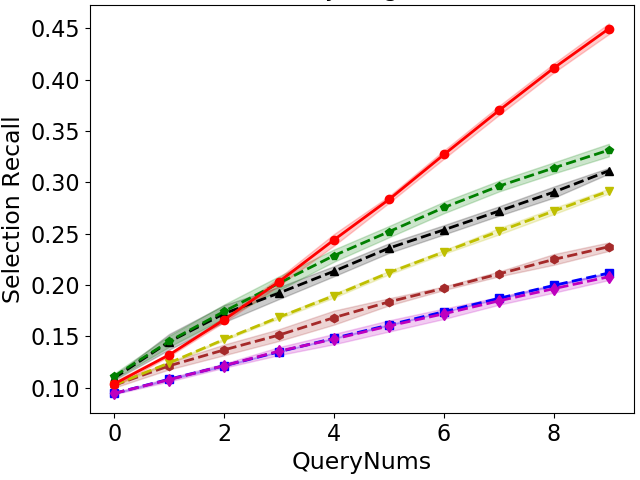}\\
		\end{subfigure}
		\begin{subfigure}{0.33\linewidth}
			\centering
			\includegraphics[width=1\textwidth]{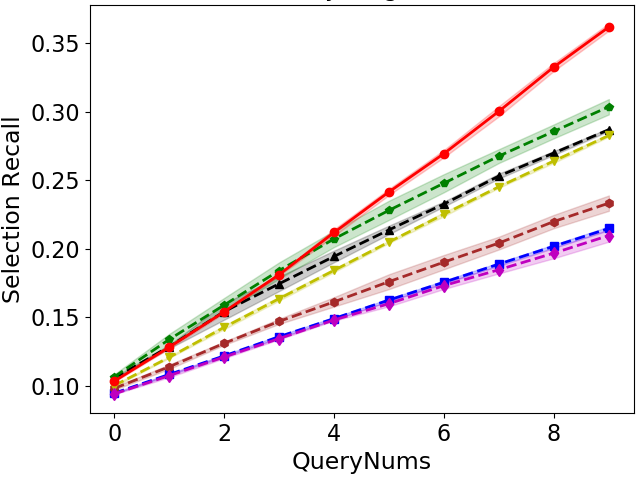}\\
		\end{subfigure}
		\begin{subfigure}{0.33\linewidth}
			\centering
			\includegraphics[width=1\textwidth]{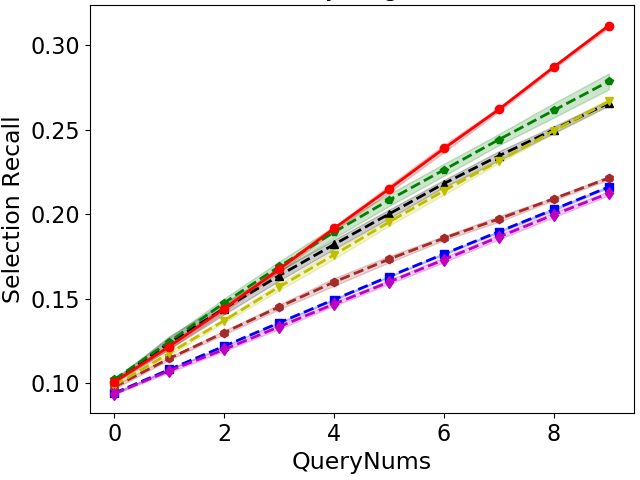}\\
		\end{subfigure}
		\caption{Selection recall comparison on CIFAR10 (first row), CIFAR100 (second row) and Tiny-Imagenet (third row) with 20\% (first column), 30\% (second column) and 40\% (third column) mismatch ratio.}
		\label{fig.comp.recall}
	\end{center}
\end{figure*}

\begin{figure*}[!ht]
	\begin{center}
		
		\begin{subfigure}{0.33\linewidth}
			\centering
			\includegraphics[width=1\textwidth]{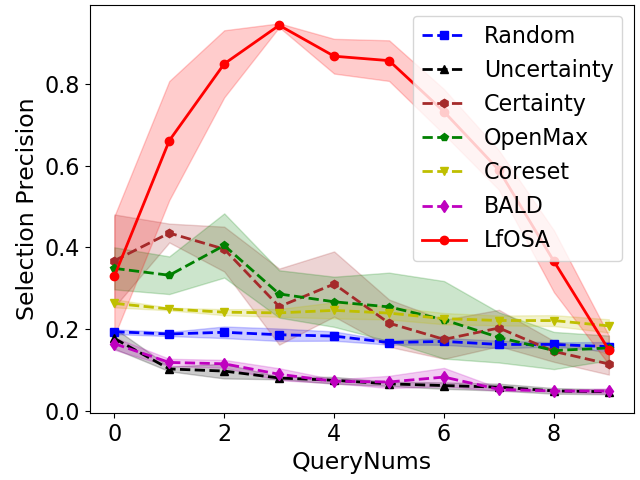}\\
		\end{subfigure}
		\begin{subfigure}{0.33\linewidth}
			\centering
			\includegraphics[width=1\textwidth]{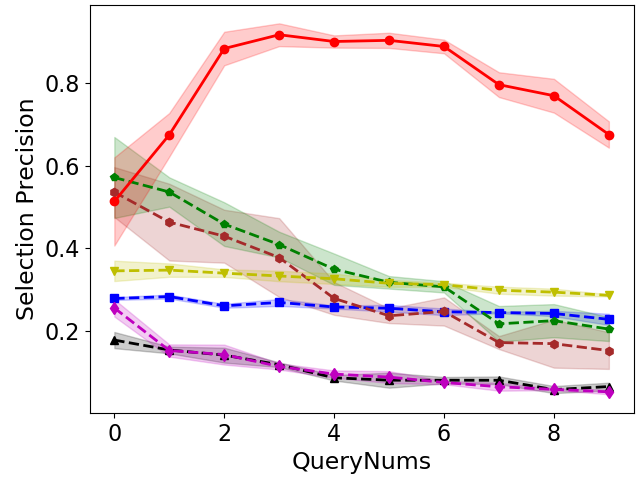}\\
		\end{subfigure}
		\begin{subfigure}{0.33\linewidth}
			\centering
			\includegraphics[width=1\textwidth]{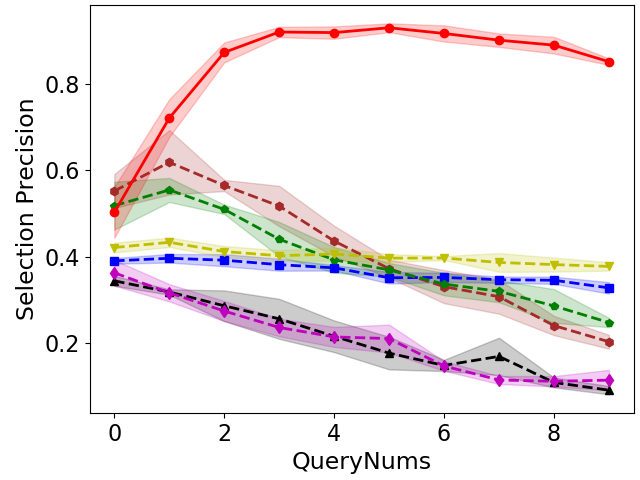}\\
		\end{subfigure}
		
		\begin{subfigure}{0.33\linewidth}
			\centering
			\includegraphics[width=1\textwidth]{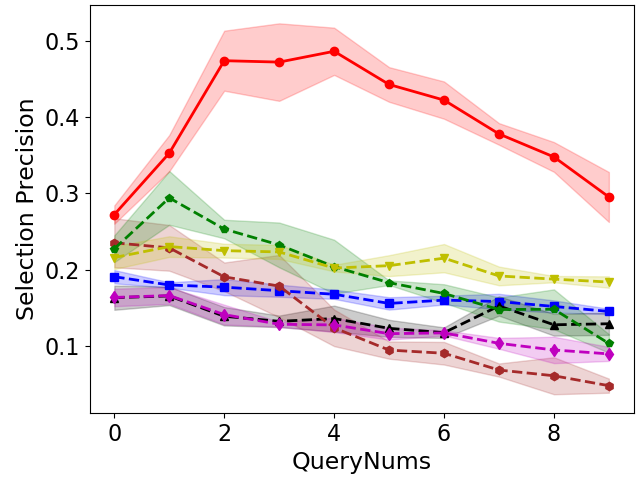}\\
		\end{subfigure}
		\begin{subfigure}{0.33\linewidth}
			\centering
			\includegraphics[width=1\textwidth]{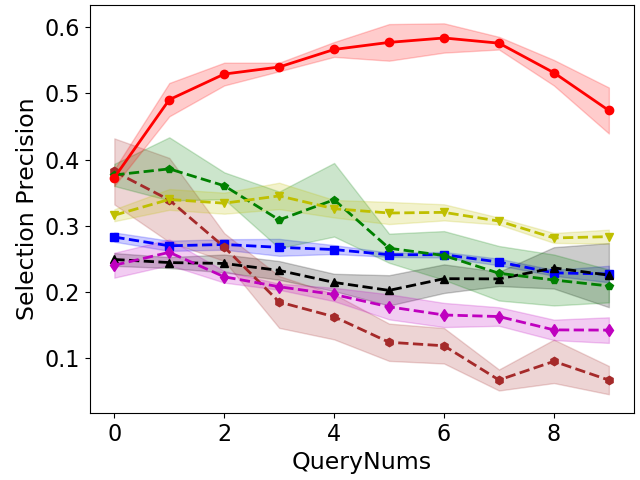}\\
		\end{subfigure}
		\begin{subfigure}{0.33\linewidth}
			\centering
			\includegraphics[width=1\textwidth]{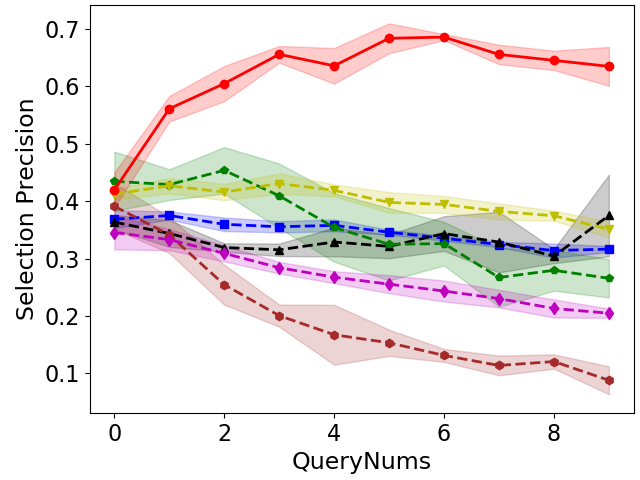}\\
		\end{subfigure}
		
		\begin{subfigure}{0.33\linewidth}
			\centering
			\includegraphics[width=1\textwidth]{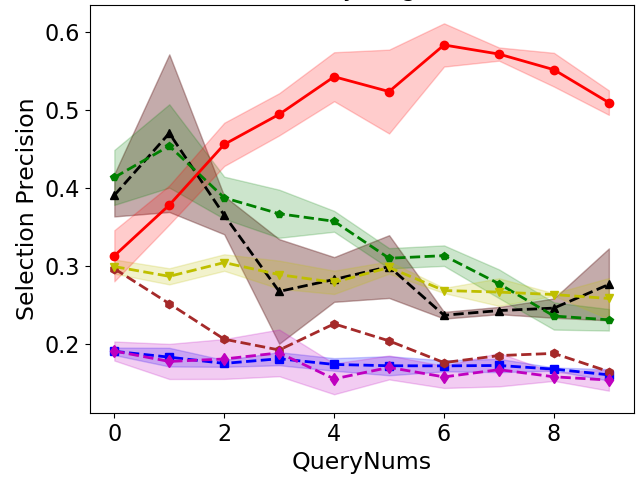}\\
		\end{subfigure}
		\begin{subfigure}{0.33\linewidth}
			\centering
			\includegraphics[width=1\textwidth]{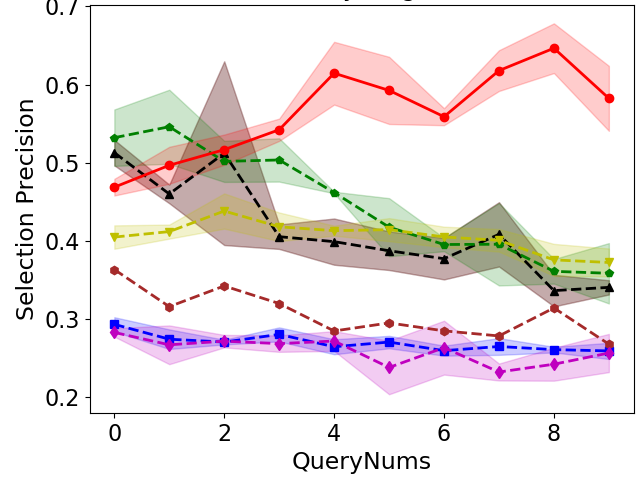}\\
		\end{subfigure}
		\begin{subfigure}{0.33\linewidth}
			\centering
			\includegraphics[width=1\textwidth]{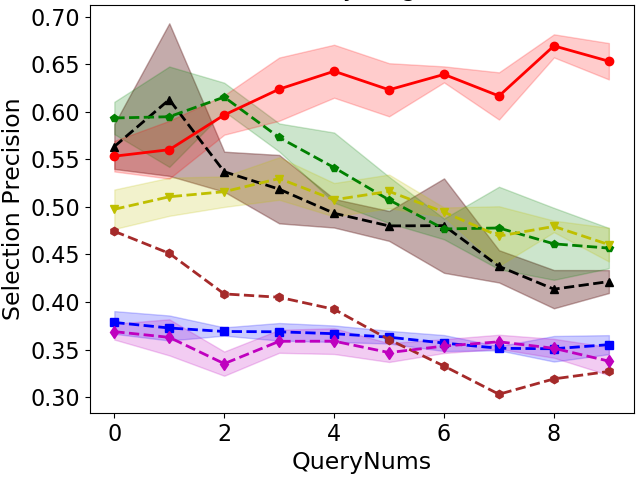}\\
		\end{subfigure}
		\caption{Selection precision comparison on CIFAR10 (first row), CIFAR100 (second row) and Tiny-Imagenet (third row) with 20\% (first column), 30\% (second column) and 40\% (third column) mismatch ratio.}
		\label{fig.comp.precision}
	\end{center}
\end{figure*}

\begin{figure*}[!ht]
	\begin{center}
		
		\begin{subfigure}{0.33\linewidth}
			\centering
			\includegraphics[width=1\textwidth]{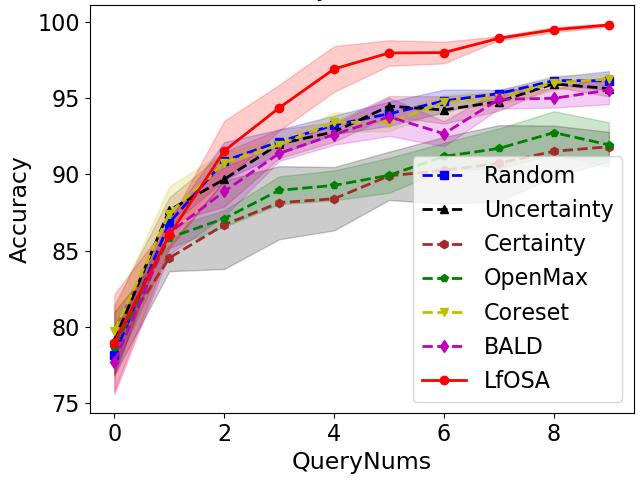}\\
		\end{subfigure}
		\begin{subfigure}{0.33\linewidth}
			\centering
			\includegraphics[width=1\textwidth]{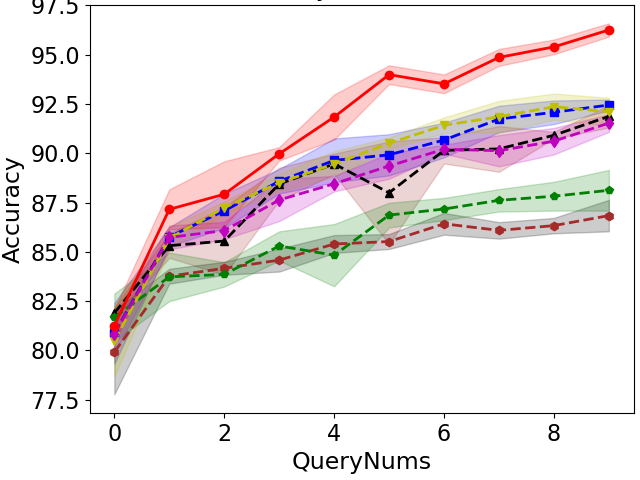}\\
		\end{subfigure}
		\begin{subfigure}{0.33\linewidth}
			\centering
			\includegraphics[width=1\textwidth]{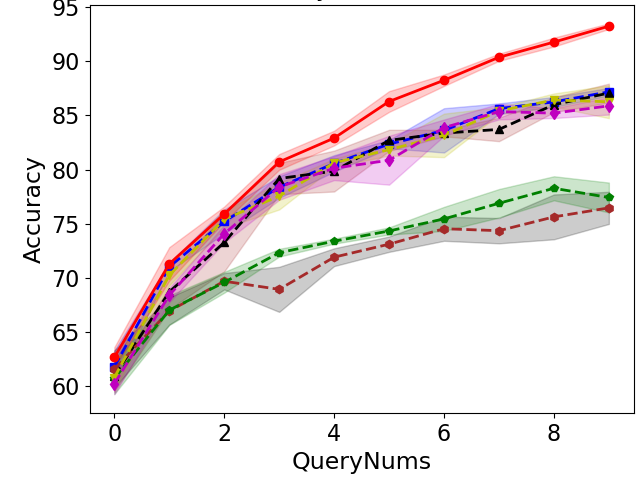}\\
		\end{subfigure}
		
		\begin{subfigure}{0.33\linewidth}
			\centering
			\includegraphics[width=1\textwidth]{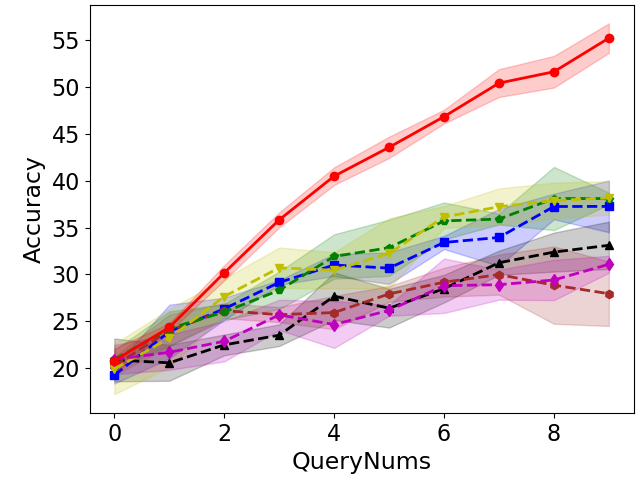}\\
		\end{subfigure}
		\begin{subfigure}{0.33\linewidth}
			\centering
			\includegraphics[width=1\textwidth]{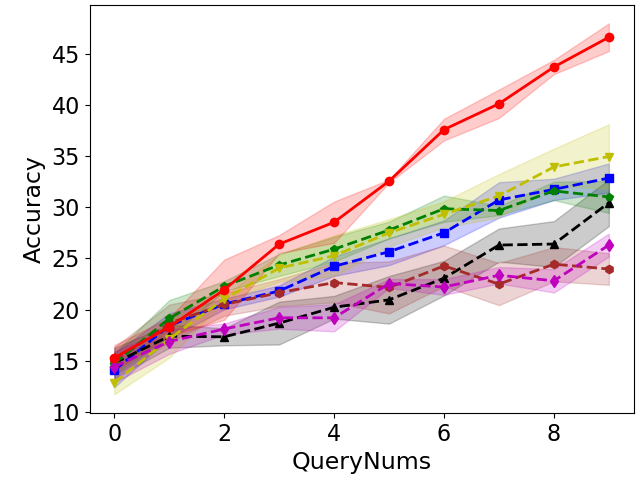}\\
		\end{subfigure}
		\begin{subfigure}{0.33\linewidth}
			\centering
			\includegraphics[width=1\textwidth]{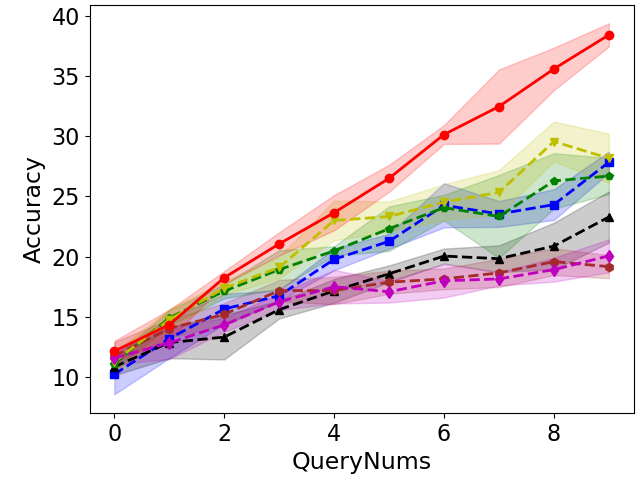}\\
		\end{subfigure}
		
		\begin{subfigure}{0.33\linewidth}
			\centering
			\includegraphics[width=1\textwidth]{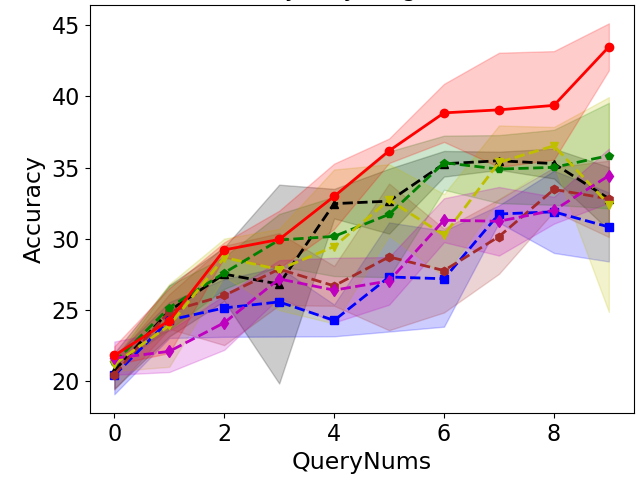}\\
		\end{subfigure}
		\begin{subfigure}{0.33\linewidth}
			\centering
			\includegraphics[width=1\textwidth]{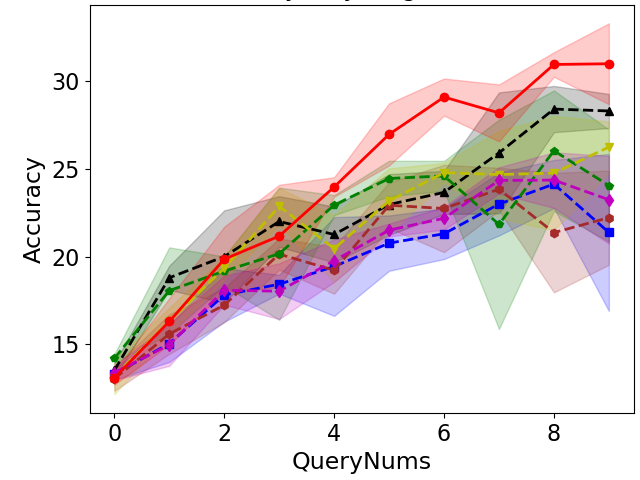}\\
		\end{subfigure}
		\begin{subfigure}{0.33\linewidth}
			\centering
			\includegraphics[width=1\textwidth]{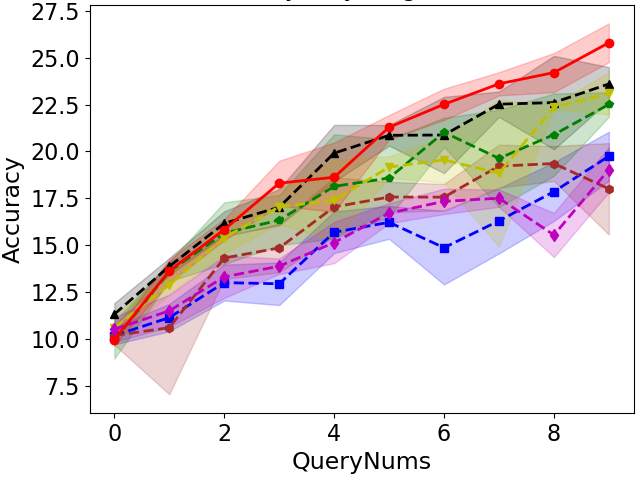}\\
		\end{subfigure}
		\caption{Classification performance comparison on CIFAR10 (first row), CIFAR100 (second row) and Tiny-Imagenet (third row) with 20\% (first column), 30\% (second column) and 40\% (third column) mismatch ratio.}
		\label{fig.comp.acc}
	\end{center}
\end{figure*}

\textbf{Classifier training.}
Based on the current labeled data $D_L$, we train the $K$-class classifier by minimizing the standard cross-entropy loss:
\begin{equation}\label{eq:loss_c}
\mathcal{L}_C(x_i, y_i) = -\sum_{i=1}^{n^L} y_i*log(f(x_i; \theta_C))
\end{equation}
where $(x_i, y_i)\in D_L$, and $n^L$ is the size of current $D_L$.

The process of the approach is summarized in Algorithm~\ref{alg:LfOSA}. Firstly, a small set of labeled data $D_L$, query batch size $b$ and temperature $T$ are given. Then the detector $\theta_D$ and classifer $\theta_C$ are randomly initialized, and the invalid set $D_I$ is initialized as an empty set. At each iteration, we train the detector by minimizing Eq.~\ref{eq:loss_r} to inference $mav_i^c$ for all unlabeled examples. Next, for each class, we collect the MAV set by the predictions of the detector and model per-example MAV to obtain known probabilities. After that, by merging and sorting these probabilities, the first $b$ examples with highest probability are selected as the query set to ask for annotation. As a result, the classifier $\theta_C$, labeled and invalid sets can be updated and output for the next iteration.

%
\section{Experiments}

To validate the effectiveness of the proposed approach, we perform experiments on CIFAR10, CIFAR100~\cite{krizhevsky2009learning} and Tiny-Imagenet~\cite{yao2015tiny} datasets, which contains 10, 100, 200 categories respectively. To construct open-set datasets, we set mismatch ratio as 20\%, 30\% and 40\% for all our experiments, where the mismatch ratio denotes the proportion of the number of known classes in the total number of classes. For example, when the mismatch ratio is set as 20\%, on CIFAR10, CIFAR100 and Tiny-Imagenet, the first 2, 20, 40 classes are known classes for classifier training, and the last 8, 80, 160 classes are seen as unknown classes respectively.

\textbf{Baselines.} To validate the effectiveness of the proposed LfOSA approach, we compare the following methods in the experiments.
\emph{i}) \textbf{Random}: it randomly selects examples from unlabeled pool for labeling.
\emph{ii}) \textbf{Uncertainty}~\cite{luo2013latent,lewis1994sequential}: it selects the examples with largest uncertainty of predictions.
\emph{iii}) \textbf{Certainty}~\cite{luo2013latent,lewis1994sequential}: it selects the examples with largest certainty of predictions.
\emph{iv}) \textbf{Coreset}~\cite{sener2017active}: it selects the representative examples by diversity.
\emph{v}) \textbf{BALD}~\cite{tran2019bayesian}: it uses dropout as an approximation to Bayesian inference for active sampling.
\emph{vi}) \textbf{OpenMax}~\cite{bendale2016towards}: a representative open-set recognition approach.
\emph{vii}) \textbf{LfOSA (ours)}: the proposed approach.

\textbf{Active learning setting.} For all AL methods, we randomly sampling 1\%, 8\% and 8\% examples as initialization labeled set on CIFAR10, CIFAR100 and Tiny-Imagenet datasets, that is, each category contains only 50, 40 and 40 examples respectively. It is worth to note that the labeled sets only contain known classes. In each AL cycle, we train a ResNet18 model for 100 epochs, SGD~\cite{zinkevich2010parallelized} is adopted as the optimizer with momentum 0.9, weight decay 5e-4, initialization learning rate 0.01, and batch size of 128, while a batch of 1500 examples is selected to query their labels for the next AL round.

\textbf{Performance measurement.} We compare the proposed LfOSA approach with other compared methods in the performance of selection recall (as Eq.~\ref{recall}), precision (as Eq.~\ref{precision}) and classification accuracy. 
Moreover, we perform the experiments for 4 runs and record the average results over 4 seeds ($seed=1,2,3,4$).

\begin{figure*}[!ht]
	\begin{center}
		
		\begin{subfigure}{0.33\linewidth}
			\centering
			\includegraphics[width=1\textwidth]{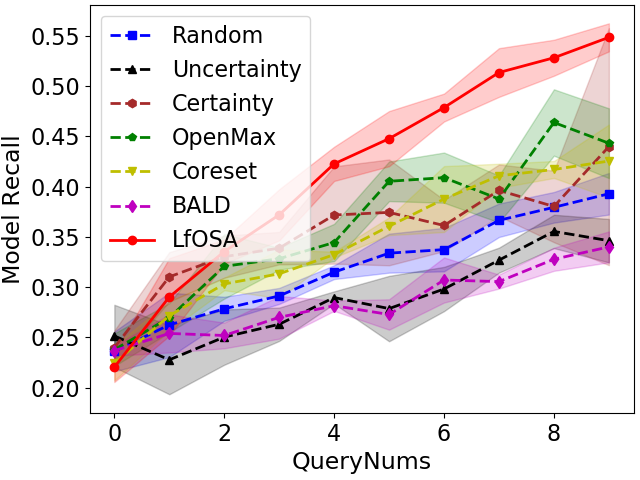}\\
		\end{subfigure}
		\begin{subfigure}{0.33\linewidth}
			\centering
			\includegraphics[width=1\textwidth]{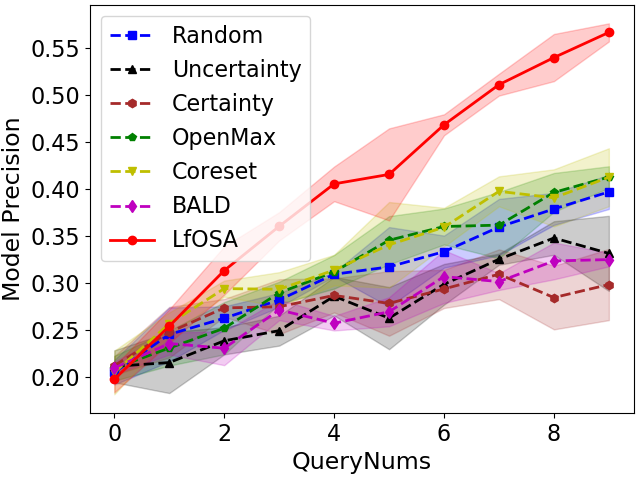}\\
		\end{subfigure}
		\begin{subfigure}{0.33\linewidth}
			\centering
			\includegraphics[width=1\textwidth]{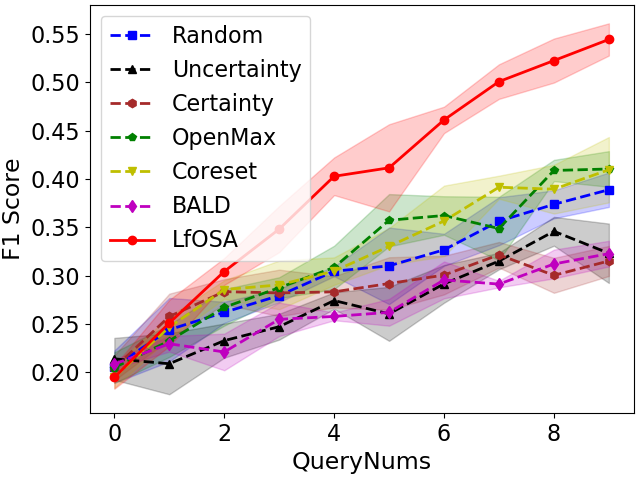}\\
		\end{subfigure}
		\caption{Classfication recall (first column), precision (second column), F1 (third column) performance comparison on CIFAR100 with 20\% mismatch ratio.}
		\label{fig.comp.f1}
	\end{center}
\end{figure*}

\subsection{Performance Comparison}
We evaluate the performance of the proposed LfOSA and compared methods by plotting curves with the number of queries increasing. The average results of recall, precision, accuracy are demonstrated in Figure~\ref{fig.comp.recall},~\ref{fig.comp.precision} and~\ref{fig.comp.acc} respectively. The first, second and third rows represent the results on CIFAR10, CIFAR100 and TinyImagenet respectively. The first, second and third columns represent the results with 20\%, 30\% and 40\% mismatch ratio.

It can be observed that no matter which dataset or mismatch ratio is used, the proposed LfOSA approach always outperforms other methods in all cases. LfOSA can achieve higher selection recall and precision during the AL process, while achieving better classification performance.
\emph{i}) For the performance of recall, the proposed LfOSA approach consistently outperforms other compared methods by a significant margin. Especially on CIFAR10 and CIFAR100, when the mismatch ratio is set to 20\%, 30\% and 40\%, the average margins between the LfOSA and Random methods are 68.8\%, 53.4\% and 35.7\% in the former and 34.3\%, 26.7\% and 20.5\% in the later.
\emph{ii}) For the performance of precision, the proposed LfOSA approach always maintain a higher selection precision than other baselines with a clear gap. It worth to note that adding invalid examples can significantly improve the detection ability (the precision of the first three queries is improving). Besides, as the number of known examples decreases, the precision is forced to decrease (the precision of the 10-th query is only 20\% on ``CIFAR10 with 20\% mismatch ratio'' because its recall has reached 96.7\%).
\emph{iii}) For the performance of classification, LfOSA consistently exhibits the best performance in all cases. Especially on CIFAR100, compared to other AL methods, LfOSA achieves about 20\%, 15\% and 12\% performance improvement under the 20\%, 30\%, 40\% mismatch ratios respectively. Moreover, with the increase of unknown ratio, the superiority of LfOSA over the other methods becomes more significant. These results indicate that the proposed LfOSA method can effectively solve the open-set annotation (OSA) problem.

\textbf{Compared methods analysis.} It is interesting to observe that two popular AL methods, Uncertainty and BALD, perform worse even than the random method in most cases. One possible reason is that these informativeness-based AL methods tend to select unknown classes, because these unknown examples are more likely to be the most informative ones. On the other hand, the Certainty method also fails in the OSA problem, which means it may not be accurate to measure the certainty of examples by using the model's prediction entropy. The diversity-based Coreset method and the open-set recognition method OpenMax show limited effectiveness in OSA tasks. The former has no recognition ability for unknown classes, and the latter lacks sufficient supervision.

\subsection{Results Using More Metrics}
To further validate the effectiveness of the proposed LfOSA approach, we compare with other methods in terms of classification recall, precision, and F1 on CIFAR100 with 20\% mismatch ratio. The experimental results are demonstrated in Figure~\ref{fig.comp.f1}.

It can be observed that the proposed LfOSA approach always significantly outperforms other methods in all cases. LfOSA can achieve higher classification recall, precision, and F1 score. These results consistently show that the proposed method can find more known examples and thus more effective training models.

\subsection{Ablation Study}
\begin{figure}[!ht]
	\begin{center}
		
		\begin{subfigure}{1\linewidth}
			\label{fig.comp.ablation.c}
			\centering
			\includegraphics[width=1\textwidth]{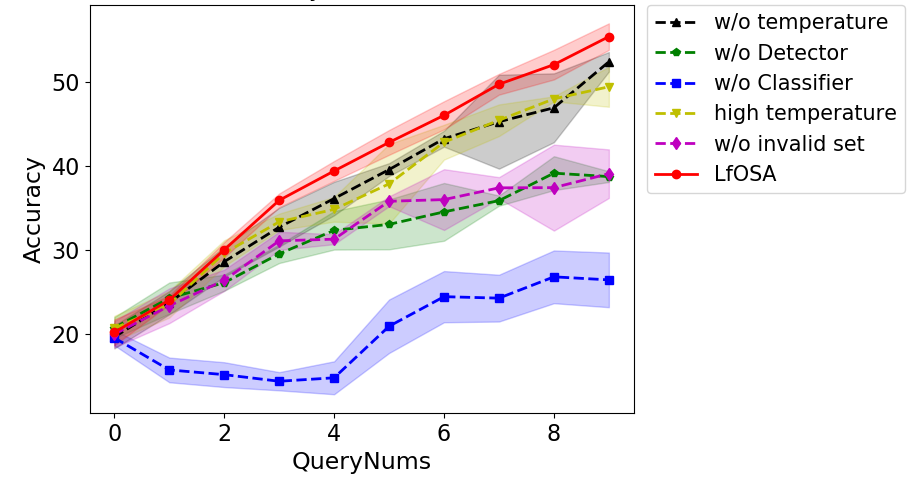}\\
		\end{subfigure}
		\caption{Ablation study on CIFAR100 with 20\% mismatch ratio.}
		\vspace{-0.4cm}
		\label{fig.comp.ablation}
	\end{center}
\end{figure}
To analyze the contribution of each component of our proposed LfOSA approach, we conduct following ablation study on CIFAR100 with 20\% mismatch ratio. The experimental results of classification accuracy are demonstrated in Figure~\ref{fig.comp.ablation}. 

\textbf{w/o temperature} and \textbf{high temperature} denote the temperature $T$ is set to 1 and 2 respectively. Compared with the LfOSA, the selection recall and classification accuracy decreased by 1.89\% and 3.1\% respectively. \textbf{w/o Detector} denotes the detector is not used, which means it employs for both detection and classification tasks. Similarly, \textbf{w/o Classifier} denote the classifier is not used. Without decoupling detection and classification, its performance is significantly deteriorated. \textbf{w/o invalid set} denotes the detector training without using the invalid set. The rapid decline of performance shows that negative examples play an essential role for detector training.

\section{Conclusion}
In this paper, we formulate a new open-set annotation (OSA) problem for real-world large-scale annotation tasks.
It introduces a practical challenge on how to maintain a high recall in identifying the examples of known classes for target model training from a massive unlabeled open-set.
To overcome this challenge, we propose an active learning framework called LfOSA to precisely select examples of known classes by decoupling detection and classification. 
By minimizing low-temperature cross-entropy loss, it exploits both known and unknown supervision to train a detector, whose activation values will be fed into a mixture Gaussian model to estimate the per-example max activation value (MAV) distribution.
Based on MAV distribution, we can distinguish examples of known classes against unknown classes in unlabeled data to build a query set for annotation.
The classifier is then updated with labeled data.
Experimental results on various tasks show the superiority of the LfOSA approach.
In the future, we will extend the OSA problem to other computer vision tasks, \eg, object detection.



{\small
\bibliographystyle{ieee_fullname}
\bibliography{cvpr22}

\begin{thebibliography}{10}\itemsep=-1pt

\bibitem{bendale2016towards}
Abhijit Bendale and Terrance~E Boult.
\newblock Towards open set deep networks.
\newblock In {\em Proceedings of the IEEE conference on computer vision and
  pattern recognition}, pages 1563--1572, 2016.

\bibitem{creswell2018generative}
Antonia Creswell, Tom White, Vincent Dumoulin, Kai Arulkumaran, Biswa Sengupta,
  and Anil~A Bharath.
\newblock Generative adversarial networks: An overview.
\newblock {\em IEEE Signal Processing Magazine}, 35(1):53--65, 2018.

\bibitem{fu2013survey}
Yifan Fu, Xingquan Zhu, and Bin Li.
\newblock A survey on instance selection for active learning.
\newblock {\em Knowledge and information systems}, 35(2):249--283, 2013.

\bibitem{ge2017generative}
ZongYuan Ge, Sergey Demyanov, Zetao Chen, and Rahil Garnavi.
\newblock Generative openmax for multi-class open set classification.
\newblock {\em arXiv preprint arXiv:1707.07418}, 2017.

\bibitem{geman1992neural}
Stuart Geman, Elie Bienenstock, and Ren{\'e} Doursat.
\newblock Neural networks and the bias/variance dilemma.
\newblock {\em Neural computation}, 4(1):1--58, 1992.

\bibitem{geng2020recent}
Chuanxing Geng, Sheng-jun Huang, and Songcan Chen.
\newblock Recent advances in open set recognition: A survey.
\newblock {\em IEEE transactions on pattern analysis and machine intelligence},
  2020.

\bibitem{goodfellow2014generative}
Ian Goodfellow, Jean Pouget-Abadie, Mehdi Mirza, Bing Xu, David Warde-Farley,
  Sherjil Ozair, Aaron Courville, and Yoshua Bengio.
\newblock Generative adversarial nets.
\newblock {\em Advances in neural information processing systems}, 27, 2014.

\bibitem{hua2008online}
Xian-Sheng Hua and Guo-Jun Qi.
\newblock Online multi-label active annotation: towards large-scale
  content-based video search.
\newblock In {\em Proceedings of the 16th ACM international conference on
  Multimedia}, pages 141--150, 2008.

\bibitem{huang2010active}
Sheng-Jun Huang, Rong Jin, and Zhi-Hua Zhou.
\newblock Active learning by querying informative and representative examples.
\newblock In {\em Advances in neural information processing systems}, pages
  892--900, 2010.

\bibitem{huang2018cost}
Sheng-Jun Huang, Jia-Wei Zhao, and Zhao-Yang Liu.
\newblock Cost-effective training of deep cnns with active model adaptation.
\newblock In {\em Proceedings of the 24th ACM SIGKDD International Conference
  on Knowledge Discovery \& Data Mining}, pages 1580--1588, 2018.

\bibitem{huang2013active}
Sheng-Jun Huang and Zhi-Hua Zhou.
\newblock Active query driven by uncertainty and diversity for incremental
  multi-label learning.
\newblock In {\em 2013 IEEE 13th International Conference on Data Mining},
  pages 1079--1084, 2013.

\bibitem{huangasynchronous}
Sheng-Jun Huang, Chen-Chen Zong, Kun-Peng Ning, and Hai-Bo Ye.
\newblock Asynchronous active learning with distributed label querying.

\bibitem{jain2014multi}
Lalit~P Jain, Walter~J Scheirer, and Terrance~E Boult.
\newblock Multi-class open set recognition using probability of inclusion.
\newblock In {\em European Conference on Computer Vision}, pages 393--409,
  2014.

\bibitem{krizhevsky2009learning}
Alex Krizhevsky, Geoffrey Hinton, et~al.
\newblock Learning multiple layers of features from tiny images.
\newblock 2009.

\bibitem{lecun2015deep}
Yann LeCun, Yoshua Bengio, and Geoffrey Hinton.
\newblock Deep learning.
\newblock {\em nature}, 521(7553):436--444, 2015.

\bibitem{lewis1994sequential}
David~D Lewis and William~A Gale.
\newblock A sequential algorithm for training text classifiers.
\newblock In {\em SIGIR’94}, pages 3--12, 1994.

\bibitem{luo2013latent}
Wenjie Luo, Alex Schwing, and Raquel Urtasun.
\newblock Latent structured active learning.
\newblock {\em Advances in Neural Information Processing Systems}, 26:728--736,
  2013.

\bibitem{ning2021improving}
Kun-Peng Ning, Lue Tao, Songcan Chen, and Sheng-Jun Huang.
\newblock Improving model robustness by adaptively correcting perturbation
  levels with active queries.
\newblock In {\em Proceedings of the AAAI Conference on Artificial
  Intelligence}, volume~35, pages 9161--9169, 2021.

\bibitem{qian2013fast}
Buyue Qian, Xiang Wang, Jun Wang, Hongfei Li, Nan Cao, Weifeng Zhi, and Ian
  Davidson.
\newblock Fast pairwise query selection for large-scale active learning to
  rank.
\newblock In {\em 2013 IEEE 13th International Conference on Data Mining},
  pages 607--616, 2013.

\bibitem{reynolds2009gaussian}
Douglas~A Reynolds.
\newblock Gaussian mixture models.
\newblock {\em Encyclopedia of biometrics}, 741:659--663, 2009.

\bibitem{roy2001toward}
N Roy and A McCallum.
\newblock Toward optimal active learning through sampling estimation of error
  reduction. int. conf. on machine learning, 2001.

\bibitem{scheirer2012toward}
Walter~J Scheirer, Anderson de Rezende~Rocha, Archana Sapkota, and Terrance~E
  Boult.
\newblock Toward open set recognition.
\newblock {\em IEEE transactions on pattern analysis and machine intelligence},
  35(7):1757--1772, 2012.

\bibitem{scheirer2014probability}
Walter~J Scheirer, Lalit~P Jain, and Terrance~E Boult.
\newblock Probability models for open set recognition.
\newblock {\em IEEE transactions on pattern analysis and machine intelligence},
  36(11):2317--2324, 2014.

\bibitem{sener2017active}
Ozan Sener and Silvio Savarese.
\newblock Active learning for convolutional neural networks: A core-set
  approach.
\newblock {\em arXiv preprint arXiv:1708.00489}, 2017.

\bibitem{settles2009active}
Burr Settles.
\newblock Active learning literature survey.
\newblock Technical report, University of Wisconsin-Madison Department of
  Computer Sciences, 2009.

\bibitem{seung1992query}
H~Sebastian Seung, Manfred Opper, and Haim Sompolinsky.
\newblock Query by committee.
\newblock In {\em Proceedings of the fifth annual workshop on Computational
  learning theory}, pages 287--294, 1992.

\bibitem{sun2010survey}
Li-Li Sun and Xi-Zhao Wang.
\newblock A survey on active learning strategy.
\newblock In {\em 2010 International Conference on Machine Learning and
  Cybernetics}, volume~1, pages 161--166, 2010.

\bibitem{tang2021qbox}
Ying-Peng Tang, Xiu-Shen Wei, Borui Zhao, and Sheng-Jun Huang.
\newblock Qbox: Partial transfer learning with active querying for object
  detection.
\newblock {\em IEEE Transactions on Neural Networks and Learning Systems},
  2021.

\bibitem{tran2019bayesian}
Toan Tran, Thanh-Toan Do, Ian Reid, and Gustavo Carneiro.
\newblock Bayesian generative active deep learning.
\newblock In {\em International Conference on Machine Learning}, pages
  6295--6304, 2019.

\bibitem{yao2015tiny}
Leon Yao and John Miller.
\newblock Tiny imagenet classification with convolutional neural networks.
\newblock {\em CS 231N}, 2(5):8, 2015.

\bibitem{you2014diverse}
Xinge You, Ruxin Wang, and Dacheng Tao.
\newblock Diverse expected gradient active learning for relative attributes.
\newblock {\em IEEE transactions on image processing}, 23(7):3203--3217, 2014.

\bibitem{zinkevich2010parallelized}
Martin Zinkevich, Markus Weimer, Alexander~J Smola, and Lihong Li.
\newblock Parallelized stochastic gradient descent.
\newblock In {\em Advances in neural information processing systems}, 2010.

\end{thebibliography}
}

\end{document}